\documentclass[runningheads]{llncs}
\usepackage{graphicx}
\usepackage{url}
\usepackage{latexsym}
\usepackage[utf8]{inputenc}
\usepackage[table]{xcolor}
\usepackage{tabularx}
\usepackage[croatian, english]{babel}
\newcolumntype{s}{>{\columncolor[HTML]{AA0044}}}
\newcommand*{\thead}[1]{\multicolumn{1}{c}{\bfseries #1}}
\begin{document}
\title{Quotations, Coreference Resolution, and Sentiment Annotations in Croatian News Articles: An Exploratory Study\thanks{Copyright © 2021 for this paper by its authors. Use permitted under Creative Commons License Attribution 4.0 International (CC BY 4.0).}}
%
%
\author{Jelena Sarajlić\inst{1}\orcidID{0000-0003-0986-3972} \and
Gaurish Thakkar\inst{1}\orcidID{0000-0002-8119-5078} \and
Diego Alves\inst{1}\orcidID{0000-0001-8311-2240} \and
Nives Mikelic Preradović\inst{1}\orcidID{0000-0001-9087-0074}}
\authorrunning{Sarajlić et al.}
%
\institute{Faculty of Humanities and Social Sciences, University of Zagreb, Zagreb 10000, Croatia \\
\email{jelenasarajlic2@gmail.com, \{dfvalio, nmikelic\}@ffzg.hr, gthakkar@m.ffzg.hr}
}
\maketitle              

\begin{abstract}
 This paper presents a corpus annotated for the task of direct-speech extraction in Croatian.
 The paper focuses on the annotation of the quotation, co-reference resolution, and sentiment annotation in SETimes news corpus in Croatian and on the analysis of its language-specific differences compared to English. From this, a list of the phenomena that require special attention when performing these annotations is derived. 
 The generated corpus with quotation features annotations can be used for multiple tasks in the field of Natural Language Processing. 

\end{abstract}

\section{Introduction}
\label{intro}
Quotes are an essential part of news articles and stories made by the media, individuals, or other organisations. The reproduction of the spoken-text includes public opinion which expresses personal and subjective information about events of the world surrounding us. Political analysts and researchers have a major interest in analysing quotations \cite{fetzer2018would} as it allows a better understanding of the political dynamics between entities as well as the identification of the correct source of claims and assertions \cite{vosoughi2018spread}.

Although recent advances in Machine Learning and Natural Language Processing do provide ample support for extracting quotes and identifying the speakers in English texts, very little research has been done for other languages. Therefore, this paper proposes an original approach for the creation of a corpus of news texts with quotes annotations for the Croatian language, with a deep analysis of the corpus genesis process. This news corpus is tagged with quotations, verb-cues, and speakers’ identification. It also includes co-reference resolution in case of pronouns involved in the quotations.  Finally, the quotes were tagged in terms of sentiment (positive, negative, or neutral) of the spoken text at the sentence level. 

This paper is organized into 6 sections. Section 2 summarizes the related work and the dataset alongside the annotation process methodology is described in Section 3. Section 4 presents several statistical information concerning the generated corpus, followed by Section 5, illustrating the various phenomena that were encountered. Finally, in Section 6, the main conclusions of this study are presented.

\section{Overview and Related Work}
The following sentences are examples of the most common syntactic arrangements of direct quotations that can be found in news articles in English.


\begin{enumerate}
    \item Michelle Mcgrath said,  ``\textit{We stand ready to support you in every way}". 
\item ``\textit{We stand ready to support you in every way}"  Blair  said. 
\item Tony Mcgrath visited Iraq... He said, “\textit{We  stand ready to support you in every way}". 
\item Tony Mcgrath visited  Iraq...  ``\textit{We stand ready to support you in every way}” the Prime Minister said.
\item ``\textit{I'm really happy for Fabio}" Materazzi told the Apcom news agency Friday. “\textit{I feel part of this distinction because I think that all the Azzurri helped a great champion like Cannavaro win an important prize}”.
\end{enumerate}

The main task of quote extraction is composed of the following sub-tasks \cite{pouliquen2007automatic}:
    Spoken-text extraction,
    Speaker identification, and 
    Verb-cue classifier.
Spoken-text extraction deals with extracting the quote's content out of the text. Speaker identification is the task of identifying the correct speaker of the extracted quote and attributing him/her to the quoted content. Verb-cue classifier must recognize the verb introducing the quote. This verb is often referred to as the `verb-cue' (for example, ``say", ``state", and other quoting verbs).
Additionally, pronouns, such as presented in case 3, must be resolved. Hence, a co-reference chain connecting the antecedent to the pronoun is needed. Furthermore, to enrich the corpus with subjective information, the text is also tagged at the sentence level in terms of sentiment (positive, negative, and neutral). 

Previous attempts of automatic extraction of quote-speaker pairs from news sources have been well-researched for English \cite{pouliquen2007automatic,krestel2008minding}. A sieve-based system in the literary text along with the dataset ``QuoteLi3" was presented by \cite{muzny2017two}. Sentiment analysis based on quotations has been extensively studied  \cite{balahur2010sentiment,balahur2009rethinking,balahur2009opinion} on English but no previous research has been done for Croatian.

\section{Corpus}
\subsection{SETIMES}
The SETimes Croatian corpus was used as the basis for the annotation process. SETimes is a parallel corpus (CC-BY-SA license) \cite{tyers2010south} based on the contents published on the SETimes.com news portal, concerning ``news and views from Southeast Europe” and covering, in total, ten languages: Bulgarian, Bosnian, Greek, English, Croatian, Macedonian, Romanian, Albanian, Serbian, and Turkish. The  Croatian corpus is composed  of 2.7 million words and 197,559 sentences.\footnote{
The corrected version of SETimes corpus where diacritics and encoding system have been corrected is available from nlp.ffzg.hr. This version is considered in this paper. 
}

\subsection{Data Pre-processing}
We merge all the sentences belonging to a single article into one document by concatenating with a space delimiter. From the whole recomposed corpus, a sub-corpus of 140 random documents was selected for the annotation task.

\subsection{Annotation}
INCEpTION tool \cite{tubiblio106270} was chosen for performing all annotations. Three custom layers were employed: 
\begin{itemize}
    \item \textbf{Quote Fine} - For tagging the various quotation components: speaker/source, verb-cue, and spoken-text. This is a span-based annotation.
    \item \textbf{Quote Simple} - For tagging the quoted text in terms of its sentiment. This is a span-based annotation which takes multiple sentences and tags them separately with the corresponding sentiment, namely positive, negative, and neutral.
    \item \textbf{Quote Co-reference} - For tagging the 3 different possible types of relations identified in quotations. 
       \begin{itemize} 
        \item \textbf{Anaphoric} - The connection of a pronoun present either in the quoted text and/or as the speaker via a co-reference chain.
        \item \textbf{Uses-verb} - Connects the speaker and verb-cue as a chain. 
        \item \textbf{Verb-spoken-text} - Connects the verb-cue to the spoken-text span as a chain.
     \end{itemize}
\end{itemize}

The annotations of the Croatian corpus were performed by a Croatian native speaker. These annotations serve as a preliminary step to try and assess the possible problems that could arise during the annotation processes and what features of quoting in Croatian should be noted when building a tool for automatic processing of quotes. 

\section{Dataset Statistics}
This section describes the overall statistics of the created dataset. We have a total of 2497 annotations concerning speaker and quote identification and types relations. In total, 469 quotes were found in 140 different documents, around 3 quotes per article. 
In terms of sentiment tagging, we have annotated 875 individual sentences.

\begin{table}[h!]
\begin{center}
\begin{tabular}{|l|r|}
\hline \bf Annotation Class & \bf No of Annotations  \\ \hline
       Speakers & 446 \\ 
       Spoken-text & 469 \\ 
       Verb-cue & 468\\ 
       Anaphoric Relation & 13\\ 
       Uses-verb Relation & 431  \\ 
       Verb-spoken-text Relation &  515\\ \hline
       Total &2497 \\
\hline
\end{tabular}
\quad
\end{center}
\caption{\label{font-table} (Left) Number of annotations in the dataset, except for sentiment tags. 
}
\end{table}
\begin{table}[h!]
\begin{center}
\begin{tabular}{|l|r|r|}
\hline \bf Class & \bf Sentences & \bf Distribution(\%) \\ \hline
      Positive & 284 & 0.32\\ 
        Neutral &  313 & 0.35\\
        Negative & 278 & 0.31\\ \hline
        Total & 875 & \\
\hline
\end{tabular}
\end{center}
\caption{\label{font-table} Sentiment distribution in the dataset. }
\end{table}
\section{Specific Croatian Quotation Linguistic Phenomena}
Quotes can be categorized into different categories, depending on which feature one would like to emphasize. In this work, the categorization stated in O'Keefe et al. \cite{o2012sequence} was used. This means differentiating between three quote types: 
\begin{itemize}
    \item \textbf{Indirect quotes} - The quoted content is not inside quotation marks and does not follow the speaker's words precisely, i.e. it is in some way changed or paraphrased.
    \item \textbf{Direct quotes} - The quoted content in its entirety is inside quotation marks and portrays the speaker's words verbatim.
    \item \textbf{Mixed quotes} - The quoted content has both direct and indirect parts, meaning that some of the speaker's words are precisely portrayed, while others are paraphrased.
\end{itemize}
The work described in this paper is focused only on direct quotes. Quote and quotation are used interchangeably, and both can indicate the whole quote (speaker+verb-cue+spoken-text) or only the quote content (`spoken-text'), depending on the context. The ideal quotation would be the one where the speaker, verb-cue, and spoken-text would be consecutively positioned in a sentence and not broken apart in any way. Often this is not the case in the corpus (nor real life), as appositions, time and date details, and other sentence parts are positioned in between the speaker, verb-cue, and spoken-text. 
To try and grasp all potential issues for automatic detection, attribution, and extraction of quotes, everything that was not deemed as the ideal quotation was written down as a problem during the manual annotation, separately from the annotations described above.
This led to many instances being marked as potential problems, but by identifying even the smallest deviations from the envisioned standard we can more easily group problems and try to adapt the automatic processing to most commonly found issues. Croatian is a South Slavic language and therefore has some distinct linguistic features when compared to English. Some problems encountered in English automatic quotation processing, such as pronoun disambiguation, do not concern Croatian (at least not to the same extent), and vice-versa -  instances which do not exist in English pose a problem in Croatian. In the following subsections, we will briefly describe all of the problems encountered during the annotation process, with a more detailed focus on language-specific ones, except for one problem because it is strictly a problem with the INCEpTION tool.
\\
 Altogether, 16 problem clusters were recognized – 15 in the Quote Fine layer and one in the Quote Simple layer. All of the problems are listed in Table 3, along with the percentage of documents\footnote{This percentage was calculated with dividing the number of documents containing a certain problem with the number of documents which had direct quotes (120 documents), since only those documents were actually annotated.} in which they were noticed. It is worth mentioning that this is merely an indicator of how many documents had a certain problem, and not how prevalent a problem might be inside a certain document (i.e. in one document the problem might arise 10 times, and only once in another document). Still, it was assessed that this could be useful as a rough estimate of problem distribution throughout the annotated documents. \\
 All of these phenomena were named by the authors for the purpose of this work (except for cross-branching and passive) and are not official names for such occurrences. The tags in examples were generated by the ReLDI tagger\footnote{ReLDI tagger, available on http://www.clarin.si/services/web/query} for Croatian.
\begin{table}[h!]    
    \begin{center}
  
    \begin{tabular}{|l|l|} 
    \hline \bf Case & \bf \% of documents \\ \hline
     Quote on multiple pages & [tool problem] \\ \hline
     Indirectly correlated speaker & 61.67\\ \hline
     Indirectly correlated verb cue & 12.50 \\ \hline
     Speaker or verb-cue in the middle of the quoted text & 41.67\\ \hline
     Quotes without an apparent speaker or verb-cue & 11.67 \\ \hline Quotes marking something other than speech & 33.33 \\ \hline
     Quotes from parts of documents or a collective & 27.50 \\ \hline Seems to be quoted but no quotes & 1.67 \\ \hline
     Speaker mentioned multiple times differently & 63.33 \\ \hline 
     Apposition(s) alongside speaker & 78.33 \\ \hline Mixed quotes & 54.17 \\ \hline Non-person speaker & 9.17 \\ \hline
     Anonymous speaker & 15.83  \\ \hline Cross-branching & 5 \\ \hline  Passive & 18.33  \\ \hline Sentence parts differently annotated & 32.50 \\ \hline
 
    \end{tabular}
    \end{center}
   \caption{\label{font-table} Problem clusters and the percentage of documents in which they were noticed.}
    \label{tab:my_label}
\end{table}

\subsection{Indirectly correlated speaker and Indirectly correlated verb-cue}
\emph{Indirectly correlated speaker} and \emph{Indirectly correlated verb-cue} were noted as two different problems as the former deals only with the speaker, and the latter only with verb-cue issues. However, they will be described together as they essentially represent the same issue, i.e. one of the other two constituents of a quote being separated from the spoken-text. This could create difficulties with automatic extraction because the speaker and verb-cue can in some cases be far away from the spoken-text. This can happen with the insertion of time and date details, appositions, or other sentence parts between the speaker, verb-cue, and spoken-text (the quoted content itself). While a human reader probably would not have any problems connecting the correct speaker or verb-cue to its spoken-text, the presumption was that this could pose a problem for the process of automatic extraction and disambiguation of quotes. While it would be possible to filter out the most common words and expressions separating the two, such as well-known appositions or time/location expressions, some less frequent instances could create a problem. This is why it was decided to note every case where the speaker or verb-cue were separated from the spoken-text, with no regard to if there was only one word separating them or whole paragraphs. This can offer an insight into how this separation happens (in our dataset) and how it could be tackled.\\ Example 6. demonstrates an occurrence of \emph{Indirectly correlated speaker}. `Sarah Lum' is the speaker whose words were quoted in the following sentence of the document, meaning that there are five tokens separating the speaker from her quote and the verb-cue. In this case, there is an apposition with additional details about the institution and its location (`predstavnica Američkog ureda u Prištini', en. `representing the US Office in Pristina') between the speaker and the quote.

\begin{table}[!ht]
\begin{center}
\begin{tabular}{c c c c c}
\thead{6. Croatian:} \\ \\
Sličnu & izjavu & dala & je & Sarah Lum,   \\
similar-Agpfsay & statement-Ncfsa & give-Vmp-sf & be-Var3s & \textbf{speaker}\\ \\
predstavnica & Američkog & ureda & u & Prištini. \\
representative-Ncfsn & American-Agpmsgy & office-Ncmsg & in-Sl & Pristina-Npfsl \\
\\``QUOTE", & kazala & je. \\
\textbf{quote} & say-Vmp-sf & be-Var3s & &\\
& \multicolumn{2}{c}{\textbf{v e r b - c u e}} & &
\end{tabular}
\end{center}
\end{table}
English: Similar comments came from Sarah Lum, representing the US office in Pristina. ``QUOTE," she said.
\\
    

In Example 7., a case of \emph{Indirectly correlated verb-cue} in a mixed quote can be seen. Verb-cue `izjavio je' (en. `has stated') is separated from the direct part of the mixed quote by the indirect part of the quote. Mixed quotes were not the focus of this work, but the direct part of a mixed quote was nevertheless annotated along with its speaker and verb-cue when judged possible or needed. It is interesting to note that there were no annotated documents which had \emph{Indirectly correlated verb-cue} where there was not also an \emph{Indirectly correlated speaker}\footnote{There was one exception, but the \emph{Indirectly correlated verb-cue} occurred in a passive clause with no agent, so there was no speaker in the first place.}, even though there were many documents with instances of \emph{Indirectly correlated speaker}, but no \emph{Indirectly correlated verb-cue}. Since these observations are made only on the document, and not on the quote-level, more detailed research would be needed to establish whether this is merely a coincidence or not.


\begin{table}[!ht]
  \begin{center}
     \begin{tabular}{c c c c c c}
     \thead{7. Croatian:}\\ \\
    ``QUOTE" & diplomatski & su & ciljevi & Ankare &\\
    \textbf{quote} & diplomatic-Agpmpny & be-Var3p & goals-Ncmpn & Ankara-Npfsg & \\
    \\ 
    na & Balkanu, & izjavio & je & Gürkan Zengin & (...) \\ 
    on-Sl & Balkan-Npmsl & state-Vmp-sm & be-Var3s & \textbf{speaker} \\
    & & \multicolumn{2}{c}{\textbf{v e r b - c u e}} & 
    \end{tabular}
    \end{center}
\end{table}

English: 
``QUOTE" are Ankara's Balkan diplomacy goals, according to \\ Gürkan Zengin (...)

\subsection{Speaker and/or verb-cue in the middle of the quoted text}
\label{Speaker and/or verb-cue in the middle of the quoted text}
This problem cluster grouped together all cases where one quote's content was separated by other text in which both the speaker and/or the verb-cue occurred. This type of quoting could be seen in Example 5. in Section 2. This meant that a quote that would be understood by a human reader as one individual quote was separated into multiple spoken-texts. This could create problems with automatic extraction of quotes because it could happen that only a part of the quote's content would be extracted with the speaker and verb-cue, while the other part could be ignored, extracted without speaker/verb-cue, or with them wrongly attributed. 
Example 8. demonstrates a quote from one Chernomorets resident, whose statement was separated by a short text with time (`krajem rujna', en. `end of September') and source details (`za SETimes', en. `for SETimes'), but also mentions speaker and verb-cue. Tags `Quote1.1' and `Quote1.2' in the example tag the first and the second part of his quote, respectively. After the second part of the quote, the article in question continues with its topic with no further reference to the fisherman or his words. Since there are no clear indications about speaker and verb-cue for the second part of the quote it could, in theory, easily be mistakenly extracted as a quote with no speaker/verb-cue (or simply ignored because of ``lack" of these constituents) or some other speaker and verb-cue could falsely be attributed to it. This is why it was thought important to note these cases so some data about how to most successfully try and add separated quote contents to each other in later steps of our work was available.

\begin{table}[!ht]
  \begin{center}
    \begin{tabular}{c c c c}
    \thead{8. Croatian:} \\ \\
      ``QUOTE1.1", & rekao & je & krajem\\
      \textbf{part 1/2 of the quote} & say-Vmp-sm & be-Var3s& end-Sg \\ 
      & \multicolumn{2}{c}{\textbf{v e r b - c u e }} & \\\\
      rujna &  za &  SETimes & Angel Kishev \\
     September-Ncmsg & for-Sa & SETimes-Npmsan & \textbf{speaker} \\ \\ 
    80-godišnji & ribar & iz & grada \\
     80-year- old-Agpmsny & fisherman-Ncmsn & from-Sg & town-Ncmsg \\ \\ 
     Chernomoretsa. & ``QUOTE1.2" & & \\
    Chernomorets-Npmsg & \textbf{part 2/2 of the quote} & & \\
    \end{tabular}
    \end{center}
\end{table}
English: ``QUOTE1.1," Angel Kishev, an 80-year-old fisherman from the town of Chernomorets, told SETimes in late September. ``QUOTE2.2"

\subsection{Quotes without an apparent speaker or verb-cue}
\label{Quotes without an apparent speaker or verb-cue}
Problem cluster \emph{Quotes without an apparent speaker or verb-cue} deals with those quotes whose speaker and/or verb-cue would be apparent to a human reader, but not necessarily to a tool for automatic quotations extraction. This could lead to quotes which are part of this cluster to be ignored or wrongly extracted/attributed. This was already partially discussed in Subsection \ref{Speaker and/or verb-cue in the middle of the quoted text}, where it was described how some parts of the separated quote could fall into this category. However, there are other cases in which this problem arises. Example 9. shows a sentence before the problematic quote. After the quote, the article continues with no additional information about its speaker or verb-cue. The sentence shown in the example contains a quote from, presumably, a different source - it seems to list the UN's endorsed standards. The real quote from Holkeri follows after that sentence. The system or the tool for automatic processing of quotes would first have to recognize that the inserted quote is not Holkeri's quote, but rather a part of some document. The next step would be to attribute Holkeri's actual quote to him, even though it does not have a specific verb-cue, is not in the same sentence as the speaker, and one has to infer that this indeed is Holkeri's quote.

\begin{table}[!ht]
  \begin{center}
    \begin{tabular}{c c c c c}
    \thead{9. Croatian:} \\ \\ 
     Plan, & dodao & je & Holkeri, & za \\
    plan-Ncmsn & add-Vmp-sm & be-Var3s & \textbf{speaker} & for-Sa\\ \\ 
    cilj & ima & osigurati & da & ciljevi \\
    goal-Ncmsan & have-Vmr3s & ensure-Vmn & that-Cs & goal-Ncmpn\\ \\
    standarda & koje & su & odredili & UN --\\
    standard-Ncmpg & which-Pi-mpa & be-Var3p & set-Vmp-pm & UN-Npmpn\\ \\
    \textbf{``}učiniti & Kosovo & boljom & sredinom & za\\
   make-Vmn & Kosovo-Npnsa & good-Agcfsiy & place-Ncfsi& for-Sa\\ \\
    sve: & sigurnom, & stabilnom & i & prosperitetnom\textbf{"} --\\
    everyone-Agpnsay & safe-Agpnsly & stable-Agpmsly & and-Cc & prosperous-Agpmsly \\ \\
    postanu & realnost. & ``QUOTE" & & \\
    become-Vmr3p & reality-Ncfsa & \textbf{quote} & & \\
    \end{tabular}
    \end{center}
    \end{table}

     English: The plan, Holkeri said, aims to ensure that the goal of the UN-endorsed standards -- ``to make Kosovo a better place for everyone: safe, stable and prosperous" -- will become a reality. ``QUOTE"

\subsection{Quotes marking something other than speech}
\label{Quotes marking something other than speech}
As the name would suggest, this problem cluster dealt with quotation marks marking something other than direct speech. These could be quotes marking informal style, quotes around some named entity etc. \emph{Quotes marking something other than speech} might ``confuse" the system or the tool for automatic processing of quotes and it could extract those non-speech instances as speech. Additionally, in some cases it is very hard to judge whether the quotation marks were used as an indication of some kind of metaphor or exaggeration, or if they actually were some speaker's literal words. This dilemma would especially be important for those dealing with all types of quotes, including mixed quotes, but it is not that crucial for direct quotation processing. Example 10. shows quotation marks which are marking a workshop's name instead of a quote.


Example 11. represents a more ambiguous case. It is unclear whether the quotation marks around `približnu istinu' (en. `approximate truth') are someone's literal words (if they are, whose?), an informal reflection of the author's or the public's feelings towards this subject or something else. Furthermore, this is an opening statement of the text (perhaps it was the article's title), and it does not get mentioned again in the text's content, making the decision even harder.


\begin{table}[!ht]
  \begin{center}
    \begin{tabular}{c c c c c}
    \thead{10. Croatian:} \\ \\
     Studenti  & (...) & organizirali & su & radionicu  \\
     student-Ncmpn &  & organize-Vmp-pm & be-Var3p & workshop-Ncfsa
    \\ \\
    ``Arhitektura, & tradicije, & sjećanje". & &\\
    architecture-Ncfsn & tradition-Ncfsg & memory-Ncnsn & & \\
    \multicolumn{3}{c}{\textbf{quotes marking something other than speech}} & &
    \end{tabular}
    \end{center}
    \vspace{-7mm}
\end{table}

English: Students (...) organised a workshop ``Architecture, Traditions, Memory".

\begin{table}[!ht]
  \begin{center}
    \begin{tabular}{c c c c}
    \thead{11. Croatian: } \\ \\
      Novi & bi & se & odbor \\
      new-Agpmsny  & be-Vaa3s & self-Px--sa &  committee-Ncmsn \\ \\
      izravno & usredotočio & na & žrtve \\
      directly-Rgp & focus-Vmp-sm & on-Sa & victim-	Ncfpa  \\ \\
     dokumentirajući & ``približnu & istinu". &\\ 
     document-Rr & approximate-Agpfsay & truth-Ncfsa & \\
     & \multicolumn{2}{c}{\textbf{quotes marking something other than speech}} &
    \end{tabular}
    \end{center}
    \vspace{-7mm}
\end{table}

English: 
    A new panel would focus squarely on the victims in documenting an ``approximation of the truth".

\subsection{Quotes from parts of documents or a collective}
\label{Quotes from parts of documents or a collective}
Quotes from a collective are not considered as ``real" quotes by the guidelines of Agence France-Presse because their real source is unclear \cite{10.1007/978-3-642-20095-3_48}. Such an outlook might not be relevant for automatic processing of quotes because one system or tool might strive for extracting all quotes, no matter if they are from a collective or an individual. However, when an article quotes some document or a part of it, this presumably wouldn't be acceptable as a quote even for the most inclusive approaches. Therefore both of these types of quotes were marked as potential problems so it could be possible to reflect on them later and decide which ones to consider and treat as quotes, and which ones to ignore. This problem will further be discussed in Subsection \ref{Passive} since it often occurred with the problem described in that subsection. 

\subsection{Seems to be quoted but no quotes}
\label{Seems to be quoted but no quotes}
\emph{Seems to be quoted, but no quotes} refers to sentences or paragraphs that seem like they could be someone's direct statement, but they are missing quotation marks. When the annotation process first started, it seemed like many documents have such occurrences. Later on, it was realized those sentences were actually source information for the article, and it was simply the reporter's name and the (presumed) title, subtitle, or introductory sentence of the article. After ignoring those introductory details of articles, only two instances of this problem were found. One such instance can be seen in Example 12., where the quote came after another quote, speaker, and verb-cue, but the quotation marks on the left side of this quote were forgotten. This could lead to the quote being missed in the automatic extraction process. 



    \begin{table}[!ht]
  \begin{center}
    \begin{tabular}{c c c c c}
    \thead{12. Croatian:} \\ \\
      Čelnici & moraju & voditi, & a & ne \\
      leader-Ncmpn & must-	Vmr3p & lead-Vmn & and-Cc & not-Qz \\ \\
      samo & pratiti & svoje &  pristaše". & \\
      only-Rgp & follow-Vmn & one's own-Px-mpa & follower-Ncmpn & \\
    \end{tabular}
    \end{center}
    \vspace{-10mm}
\end{table}

English: Leaders must lead, and not merely follow their followers".



\subsection{Speaker mentioned multiple times differently}
\label{Speaker mentioned multiple times differently}
In news articles, one individual speaker is often referred to in many ways after he was first introduced to the text. In English, this might be done with the surname of the speaker, his title (such as ``prime minister") or a pronoun, which then creates a problem because pronoun disambiguation is needed. In Croatian, it is not usual to use pronouns when referring to a person without using their name/title because the predicate form often indicates gender, so pronoun disambiguation was not something crucial to have in mind at this stage. In the rare article that this was the case, the pronoun was annotated as having an anaphoric relationship with the name. Surnames and titles are also often used to refer to the speaker. Creating some sort of connections between surnames/titles and the speaker's full name will be needed in the future to identify the speaker's full name for all his or her quotes found in the text. In general, different versions of someone's name were not annotated as anaphoric (so, surnames were not annotated as having an anaphoric relationship with the prename or full name). When the speaker of the quote was the speaker's title, the title and the name were tagged as having an anaphoric relationship. 

\subsection{Apposition(s) alongside speaker}
\label{Apposition(s) alongside speaker}
Many times when a source is quoted, his or her title or some other description of who this person is (like the `80-year-old fisherman' in Example 8.) are also mentioned. All of these occurrences were grouped together under the problem cluster \emph{Apposition(s) alongside speaker} and marked when an apposition of any kind would appear next to the speaker. This problem cluster is not exactly a problem because appositions alongside speakers would not in any way intrude the process of automatic extraction/disambiguation of quotations. On the contrary, they were marked to gain a rough overview of what kinds of expressions could be used instead of the speaker's name when referring to him/her and how to extract and connect them to the speaker's name. The potential dataset one could gather from a collection of appositions could be used for resolving issues described in Subsection \ref{Speaker mentioned multiple times differently}.
\subsection{Mixed quotes}
Mixed quotes are quotes that combine direct and indirect quoting styles. While this work is currently focused only on direct quotations, it was strived to annotate direct parts of mixed quotes whenever it was judged possible or necessary to do so, most often when the meaning or purpose of the direct part would be clear enough on its own. Example 13. demonstrates a mixed quote - parts of the speaker's statement were put in quotation marks, and other parts (the speaker expressing condolences) were indirectly quoted. The mixed quote in this example is even more problematic than others could be because the directly quoted statement is expressed in the 3\textsuperscript{rd} person singular, and not in 1\textsuperscript{st} person singular as one would expect of someone speaking for himself. Based on that, it can be concluded that even this direct part was changed and possibly filtered by the article's author.


    \begin{table}[!h]
  \begin{center}
    \begin{tabular}{c c c c c c}
    \thead{13. Croatian:}\\
    Rekavši & kako & se & ``sjeća, & te & kako \\
    say-Rr & how-Cs & self-Px--sa & remember-Vmr3s & and-Cc & how-Cs\\ \\
    je & svjestan & dubine & (...)" & Peres & je\\
    be-Var3s & aware-Agpmsnn & depth-Ncfsg & & \textbf{speaker} & be-Var3s\\ \\
   izrazio & sućut & (...)\\
    express-Vmp-sm & condolence-Ncfsa & &  
    \end{tabular}
    \end{center}
\end{table}

    English: Saying he ``remembers and aware of the depth (...)," Peres extended condolences (...) \\ \\
Direct parts of mixed quotes sometimes do not have their own verb-cue because the verb-cue with which they were introduced is not a typical quoting verb. Example 14. portrays this nicely, as `izrazio žaljenje' (en. `voiced regret') is far from what one would consider a typical quoting verb. Often these annotated direct parts of mixed quotes were annotated only partially - without verb-cue or without the connections in the Quote Co-reference layer.

\vspace{-4 mm}
\begin{table}[!ht]
  \begin{center}

    \begin{tabular}{c c c c c}
    \thead{14. Croatian:}  \\ \\
    
      (...)   & Holkeri &  je & također & izrazio \\
         & \textbf{speaker} & be-Var3s & also-Rgp & express-Vmp-sm \\ \\
         žaljenje & jer & ``nisu & sve & zajednice"  \\
        regret-Ncnsa & because-Cs &be-Var3p & all-Agpfpny & community-Ncfpn\\ \\ 
         sudjelovale & u & izradi & plana. & \\
         participate-Vmp-pf & in-Sl & development-Ncfsl & plan-Ncmsg &
    \end{tabular}\\

    \end{center}
\end{table}
\vspace{-4 mm}

    English:
    (...) Holkeri also voiced regret over the fact that ``not every community" participated in the development of the plan.
\vspace{-4.2mm}

\subsection{Non-person speaker and Anonymous speaker}
These two problems were marked down separately, but both concern issues with speaker disambiguation and attribution so they will briefly be described together.\\ \emph{Non-person speaker} was used to mark those sources which were not persons, but rather collectives or documents. Because it was often found alongside \emph{Quotes from parts of documents or a collective}, it was used as an addition to that problem cluster. \\
\emph{Anonymous speaker} problem cluster marked all speakers who were not mentioned by their name, but rather with an apposition, title etc. Example 15. has a quote whose source is a collective - `dvoje čelnika' (en. `two leaders') - and additionally, they are not named. Future work should decide on how to treat such ``speakers" and their quotes, as already mentioned in Subsection \ref{Quotes from parts of documents or a collective}.

\vspace{-3mm}
\begin{table}[!ht]
  \begin{center}
    \begin{tabular}{c c c c c}
    \thead{15. Croatian:}\\ \\
    ``QUOTE", & kazalo & je & dvoje & čelnika \\
    \textbf{quote} & say-Vap-sn & be-Var3s & two-Mls & leader-Ncmpg \\ 
     u & priopćenju & nakon & sastanka. & \\
     in-Sl & statement-Ncnsl & after-Sg & meeting-Ncmsg & 
    \end{tabular}
    \end{center}
    \vspace{-8mm}
\end{table}

English: ``QUOTE" the two leaders said in a statement after the meeting.

\subsection{Cross-branching} 
\label{Cross-branching}
This phenomenon is also referred to as ``discontinuous constituents"\footnote{The term cross-branching is used in 
\cite{lai2013anatomy}, while the term discontinuous constituents can be found in 
\cite{van2001introduction}. In 
\cite{volk2007search} crossing edges are mentioned. Other terms could also be used to describe this phenomenon.} and occurs when some sentence parts split other sentence parts, such as predicates or noun groups, by appearing in the middle of the second sentence part's construction. In Croatian, it is often observed as a splitting of the predicate.

As evident in Example 16., the auxiliary part of the predicate is separated from the participle part by four other tokens (in the example, the notation P means predicate).
This is a problem for verb cue classifiers because they would have to recognize such instances and filter out the full predicate from the sentence. However, this could be easily solved by automatically adding the auxiliary verb/copula when what seems like a lone participle is found. The gender of the auxiliary verb can be easily deduced from the morphological form of the participle, since parts of a predicate must agree in gender.

\vspace{-4 mm}
\begin{table}[!ht]
\begin{center}
\begin{tabular}{c c c}
\thead{16. Croatian:} \\ \\
     Vojnici & su & svoje  \\ 
     soldier-Ncmpn & be-Var3p & their-own-Px-nsa\\
     & \textbf{aux. part of the P} &  \\ \\
    angažiranje & u & Iraku\\  
     engagement-Ncnsa & in-Sl & Iraq-Npmsl \\ \\ 
      opisali & kao & "QUOTE" \\
     describe-Vmp-pm & as-Cs & \textbf{quote} \\
     \textbf{participle part of the P} & &
\end{tabular}
    \end{center}
\vspace{-10mm}
\end{table}

 English:
The soldiers described the engagement in Iraq as ``QUOTE"
\subsection{Passive}
\label{Passive}
Passive in Croatian usually has no agent, meaning that the quotes whose verb-cue is in passive would be without a speaker. The Croatian school grammar states that ``passive is used when the agent is unknown or one doesn't wish to specifically emphasize the agent" \cite{hudevcek2017hrvatska}. As one could expect, in all of the documents in which \emph{Passive} was noted, \emph{Quotes from parts of documents or a collective} was also noted. This continues the problem discussed in Subsection \ref{Quotes from parts of documents or a collective}, i.e. whether those quotes should really be thought of as quotes or not. Sentences with a predicate in passive are also sometimes vague in the sense of the quote's source, like in Example 17.
\noindent
\vspace{-4mm}
\begin{table}[!ht]
  \begin{center}
\begin{tabular}{c c c c c}
\thead{17. Croatian:}\\
     ``QUOTE", &  navodi & se & u & priopćenju. \\
     \textbf{quote} & state-Vmr3s & self-Px--sa & in-	Sl & statement-Ncnsl \\ 
     & \multicolumn{2}{c}{\textbf{p  a  s  s  i  v  e}}
    \end{tabular}
    \end{center}
    \vspace{-12mm}%
\end{table}
English: ``QUOTE", it is stated in the statement.

\subsection{Sentence parts differently annotated}
  Some of the annotated quotes presented more than one sentiment due to a change of tone or topic. Such sentences were segmented into parts which then received different sentiment annotations. These types of sentences can usually be easily spotted in Croatian because they use so-called ``contrary conjunctions" which link constituent-sentences of contradictory sentiments. Contrary sentences are a type of complex sentences in which all of the clauses can be independent and convey contrary meanings. An example can be seen in Example 18., where `ali' (en. `but') is the contrary conjunction. The part before `ali' was annotated as positive and the rest of the sentence as negative. For easier understanding, the positive clause has been coloured with green, while the negative one was coloured with orange in the example given.



\begin{table}[!ht]
  \begin{center}
    \begin{tabular}{c c c c c}
        \thead{18. Croatian:} \\
      \rowcolor[HTML]{C1FAAC}``Vjernik & sam & i & ponosim & se \\
      believer-Ncmsn & be-Var1s & and-Cc & proud-	Vmr1s & self-Px--sa\\ 
      \multicolumn{5}{c}{\textbf{p  o  s  i  t  i  v  e}} \\ \\ 
       \cellcolor[HTML]{C1FAAC}time, & ali & \cellcolor[HTML]{FAD6AC}ubijanje & \cellcolor[HTML]{FAD6AC}nevinih & \cellcolor[HTML]{FAD6AC}ljudi\\ \\
       it-Pd-nsi & but-Cc & killing-Ncnsn & innocent-Agpmpgy & people-Ncmpg \\ 
       & \textbf{contrary conjuction} & & & \\ \\ 
     \rowcolor[HTML]{FAD6AC} ne &  može & nikako & biti & (...)" \\
      no-Qz & can-Vmr3s & no way-Rgp & be-Van \\
       \multicolumn{5}{c}{\textbf{n e g a t i v e}}
    \end{tabular}
    \vspace{-10mm}%
    \end{center}
\end{table}

\begin{enumerate}
 \item[] English: ``I am faithful and proud of it, \&  but killing innocent people can not have (...)"
\end{enumerate}

\section{Conclusions}
In our paper, a detailed presentation of how a Croatian corpus of 140 documents annotated in terms of direct quotation features was created and what procedures were employed is offered. Potential problems for future work in the Quote Fine and Quote Simple layer were recognized and described, along with their examples in Croatian and English. The work described in this paper provides merely a starting point in creating a system or a tool for automatic extraction and attribution of quotations in Croatian. In the future, this gold standard data will be used to automatically annotate the remaining un-tagged documents of Croatian SETimes collection and create a silver standard dataset for Croatian. Furthermore, the annotations will be projected to the SETimes English parallel corpus and Bulgarian, Bosnian, Greek, Macedonian, Romanian, Albanian, Serbian, and Turkish.

\section{Acknowledgements}
The work presented in this paper has received funding from the European Union’s Horizon 2020 research and innovation program under the Marie Skłodowska-Curie grant agreement no. 812997 and under the name CLEOPATRA (Cross-lingual Event-centric Open Analytics Research Academy).
\bibliographystyle{splncs04}
\bibliography{mybibliography}

\begin{thebibliography}{10}
\providecommand{\url}[1]{\texttt{#1}}
\providecommand{\urlprefix}{URL }
\providecommand{\doi}[1]{https://doi.org/#1}

\bibitem{balahur2009rethinking}
Balahur, A., Steinberger, R.: Rethinking sentiment analysis in the news: from
  theory to practice and back  (2009)

\bibitem{balahur2010sentiment}
Balahur, A., Steinberger, R., Kabadjov, M., Zavarella, V., van~der Goot, E.,
  Halkia, M., Pouliquen, B., Belyaeva, J.: Sentiment analysis in the news. In:
  Proceedings of the Seventh International Conference on Language Resources and
  Evaluation (LREC'10) (2010)

\bibitem{balahur2009opinion}
Balahur, A., Steinberger, R., Van Der~Goot, E., Pouliquen, B., Kabadjov, M.:
  Opinion mining on newspaper quotations. In: 2009 IEEE/WIC/ACM International
  Joint Conference on Web Intelligence and Intelligent Agent Technology.
  vol.~3, pp. 523--526. IEEE (2009)

\bibitem{fetzer2018would}
Fetzer, A., Weizman, E.: ‘what i would say to john and everyone like john
  is...’: The construction of ordinariness through quotations in mediated
  political discourse. Discourse \& Society  \textbf{29}(5),  495--513 (2018)

\bibitem{hudevcek2017hrvatska}
Hude{\v{c}}ek, L., Mihaljevi{\'c}, M.: Hrvatska {\v{s}}kolska gramatika  (2017)

\bibitem{tubiblio106270}
Klie, J.C., Bugert, M., Boullosa, B., de~Castilho, R.E., Gurevych, I.: The
  inception platform: Machine-assisted and knowledge-oriented interactive
  annotation. In: Proceedings of the 27th International Conference on
  Computational Linguistics: System Demonstrations. pp.~5--9. Association for
  Computational Linguistics (June 2018),
  \url{http://tubiblio.ulb.tu-darmstadt.de/106270/}

\bibitem{krestel2008minding}
Krestel, R., Bergler, S., Witte, R., et~al.: Minding the source: Automatic
  tagging of reported speech in newspaper articles. Reporter  \textbf{1}(5), ~4
  (2008)

\bibitem{10.1007/978-3-642-20095-3_48}
de~La~Clergerie, {\'E}., Sagot, B., Stern, R., Denis, P., Recourc{\'e}, G.,
  Mignot, V.: Extracting and visualizing quotations from news wires. In:
  Vetulani, Z. (ed.) Human Language Technology. Challenges for Computer Science
  and Linguistics. pp. 522--532. Springer Berlin Heidelberg, Berlin, Heidelberg
  (2011)

\bibitem{lai2013anatomy}
Lai, P.Y.: The anatomy of translation problems. Chartridge Books Oxford (2013)

\bibitem{muzny2017two}
Muzny, G., Fang, M., Chang, A., Jurafsky, D.: A two-stage sieve approach for
  quote attribution. In: Proceedings of the 15th Conference of the European
  Chapter of the Association for Computational Linguistics: Volume 1, Long
  Papers. pp. 460--470 (2017)

\bibitem{o2012sequence}
O’Keefe, T., Pareti, S., Curran, J.R., Koprinska, I., Honnibal, M.: A
  sequence labelling approach to quote attribution. In: Proceedings of the 2012
  Joint Conference on Empirical Methods in Natural Language Processing and
  Computational Natural Language Learning. pp. 790--799 (2012)

\bibitem{pouliquen2007automatic}
Pouliquen, B., Steinberger, R., Best, C.: Automatic detection of quotations in
  multilingual news. In: Proceedings of Recent Advances in Natural Language
  Processing. pp. 487--492 (2007)

\bibitem{tyers2010south}
Tyers, F.M., Alperen, M.S.: South-east european times: A parallel corpus of
  balkan languages. In: Proceedings of the LREC Workshop on Exploitation of
  Multilingual Resources and Tools for Central and (South-) Eastern European
  Languages. pp. 49--53 (2010)

\bibitem{van2001introduction}
Van Valin~Jr, R.D., et~al.: An introduction to syntax. Cambridge University
  Press (2001)

\bibitem{volk2007search}
Volk, M., Lundborg, J., Mettler, M.: A search tool for parallel treebanks
  (2007)

\bibitem{vosoughi2018spread}
Vosoughi, S., Roy, D., Aral, S.: The spread of true and false news online.
  Science  \textbf{359}(6380),  1146--1151 (2018)

\end{thebibliography}




\end{document}